\documentclass[]{bytedance}
\usepackage[toc,page,header]{appendix}
\usepackage{minitoc}
\usepackage{algorithm}
\usepackage{algpseudocode}
\usepackage{algpseudocode}
\usepackage{booktabs}
\usepackage{multirow}
\usepackage{siunitx}
\usepackage{makecell}
\usepackage{mathtools}
\usepackage{tabularx,array}
\usepackage[table]{xcolor}
\usepackage{graphicx}
\usepackage{wrapfig}
\usepackage{xspace}
\usepackage{amsmath}
\usepackage{amsfonts}
\usepackage{caption}

\usepackage{epsfig}
\usepackage{amssymb}
\usepackage{etoolbox}
\usepackage{listings}
\usepackage{enumitem}
\usepackage{amsthm}

\usepackage{hyperref}
\usepackage{xcolor}

\title{\method: Towards \underline{Hi}gh-\underline{Fi}delity Reference-Based \underline{Inpaint}ing for Generating Detail-Preserving Human-Product Images}

\newcommand{\ie}{\textit{i.e.}}

\newcommand{\method}{HiFi-Inpaint\xspace}

\newcommand{\myparagraph}[1]{\noindent\textbf{#1}}

\author[1,\star]{Yichen Liu}
\author[2,\star]{Donghao Zhou}
\author[3]{Jie Wang}
\author[3]{Xin Gao}
\author[3]{Guisheng Liu}
\author[3,\dagger]{\\Jiatong Li}
\author[4]{Quanwei Zhang}
\author[1]{Qiang Lyu}
\author[5]{Lanqing Guo}
\author[3,\S]{\\Shilei Wen}
\author[1,\S]{Weiqiang Wang}
\author[2,\S]{Pheng-Ann Heng}

\affiliation[1]{University of Chinese Academy of Sciences}
\affiliation[2]{The Chinese University of Hong Kong}
\affiliation[3]{ByteDance}
\affiliation[4]{Zhejiang University}
\affiliation[5]{UT Austin}

\contribution[\star]{Equal contribution}
\contribution[\dagger]{Project lead}
\contribution[\S]{Corresponding author}

\abstract{
Human-product images, which showcase the integration of humans and products, play a vital role in advertising, e-commerce, and digital marketing.
The essential challenge of generating such images lies in \textit{ensuring the high-fidelity preservation of product details}.
Among existing paradigms, \textit{reference-based inpainting} offers a targeted solution by leveraging product reference images to guide the inpainting process. 
However, limitations remain in three key aspects: 
the lack of diverse large-scale training data,
the struggle of current models to focus on product detail preservation, and
the inability of coarse supervision for achieving precise guidance.
To address these issues, we propose \textbf{\method}, a novel high-fidelity reference-based inpainting framework tailored for generating human-product images. 
\method introduces \textit{Shared Enhancement Attention (SEA)} to refine fine-grained product features and \textit{Detail-Aware Loss (DAL)} to enforce precise pixel-level supervision using high-frequency maps. 
Additionally, we construct a new dataset, \textit{HP-Image-40K}, with samples curated from self-synthesis data and processed with automatic filtering.  
Experimental results show that \method achieves state-of-the-art performance, delivering detail-preserving human-product images.
}

\checkdata[Note]{Accepted by CVPR 2026}
\checkdata[Project Page]{\url{https://correr-zhou.github.io/HiFi-Inpaint}}

\begin{document}
\maketitle

\section{Introduction}
\label{sec:intro}

% task
Recent advancements in diffusion models \cite{SD,peebles2023scalable,FLUX} have revolutionized image generation, enabling the creation of diverse visual content across real-world scenarios \cite{cao2024controllable,huang2025diffusion,wei2025personalized}.
Among these, generating human-product images has emerged as a practical application in industries such as advertising, e-commerce, and digital marketing.
This task takes textual descriptions and visual cues as inputs, generating images that seamlessly integrate humans with products.
By creating high-quality human-product images, this task enables the automated production of commercial content at unprecedented speed and quality, unlocking new horizons for the real-world impact of image generation.
Importantly, it reduces manual design effort while improving image quality and consistency across large-scale deployments.

However, generating high-quality human-product images poses significant challenges
and the most critical one lies in \textit{ensuring the high-fidelity preservation of product details}. 
Specifically, generated human-product mages should faithfully depict the fine-grained features of products, such as shapes, colors, patterns, and texture, to meet the stringent requirements of real-world use.
Otherwise, even subtle inaccuracies can undermine consumer trust and reduce the effectiveness of commercial efforts.
Utilizing existing paradigms, such as image customization \cite{ruiz2023dreambooth} or text-driven editing \cite{chen2025empirical}, to generate these highly demanding images often leads to inferior results. 
This is because they typically focus on global or high-level semantic manipulation with a free-form input manner, making it difficult to robustly preserve fine-grained details.

One promising paradigm is \textit{reference-based inpainting} \cite{yang2023paint}, which synthesizes coherent images by leveraging reference images to guide the inpainting process. This approach offers a more structured and targeted mechanism for integrating product-specific features while maintaining overall consistency.
However, existing methods \cite{song2025insert,mao2025ace++,li2025ic} still fall short in preserving fine-grained product details. This is largely because diffusion-based methods struggle to faithfully retain reference image details due to limited enforcement of strict spatial and appearance alignment. The denoising process tends to average or hallucinate content, resulting in inconsistencies in texture, shape, and branding elements that are critical for high-fidelity demands.

\begin{figure}
    \centering
    
    % \vspace{-7mm}
    
    \includegraphics[width=1\linewidth]{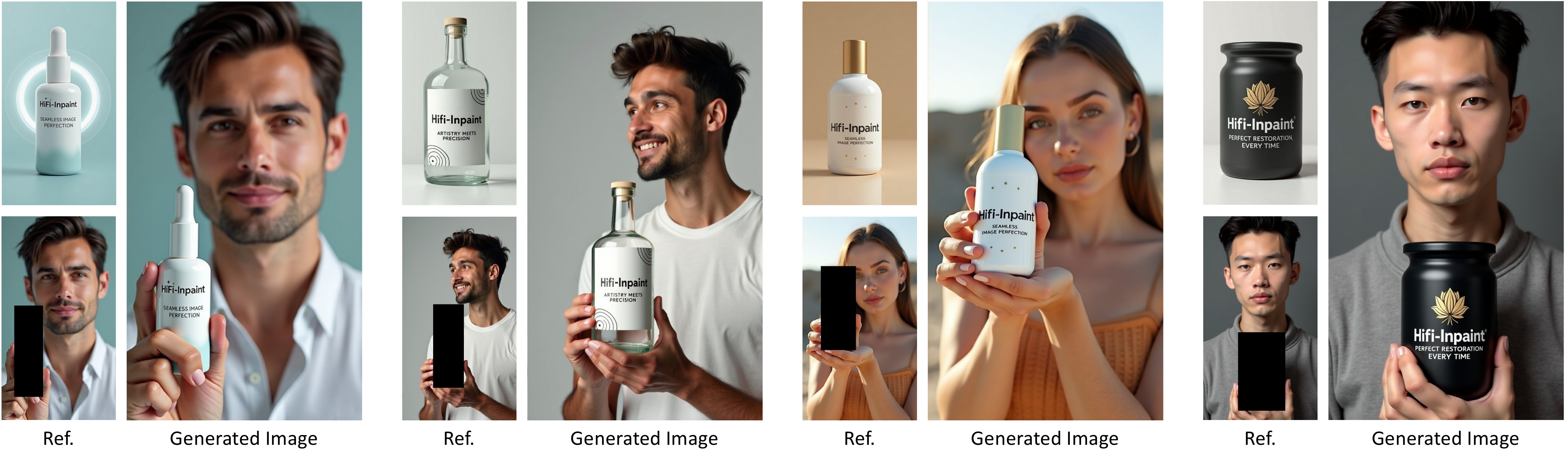}
    % \includegraphics[width=0.7\linewidth]{example-image-golden}
    % \rule{1\linewidth}{0.5\linewidth}
    
    \vspace{-1mm}
    
    \captionsetup{hypcap=false}
    \captionof{figure}
    { 
        \textbf{\method enables high-fidelity reference-based inpainting.}
        Our \method can seamlessly integrate product reference images into masked human images, generating high-quality human-product images with high-fidelity detail preservation.
        To avoid potential privacy and copyright concerns, we use AI-generated products and humans for presentation purposes in this paper.
        \textit{Zoom in for better view.}
    }
    \vspace{-2mm}
    \label{fig:teaser}
    
\end{figure}

To address these limitations, we propose \textbf{\method}, a novel \underline{\textbf{Hi}}gh-\underline{\textbf{Fi}}delity reference-based \underline{\textbf{Inpaint}}ing framework tailored for generating detail-preserving human-product images (Fig.~\ref{fig:pipeline}).
Our \method produces visually consistent and appealing human-product images using a concise text prompt, a masked human image, and a product reference image as inputs.
First, to support robust model training, we construct a new dataset, \textbf{HP-Image-40K}, consisting of 40,000+ samples generated via a self-synthesis pipeline followed by automatic filtering to ensure high-quality, diverse training data.
We then design a high-frequency map-guided DiT framework that employs a token merging mechanism to effectively coordinate the integration of multiple image conditions, which additionally considers the injection of high-frequency maps.
To further enhance visual fidelity, we introduce \textbf{Shared Enhancement Attention (SEA)}, which utilizes shared dual-stream visual DiT blocks to refine visual tokens within masked regions. By replacing product image tokens with corresponding high-frequency map tokens in another branch, SEA helps to enhance fine-grained product features for the original branch of the base model.
Finally, we propose a \textbf{Detail-Aware Loss (DAL)} that leverages high-frequency pixel-level supervision to guide the reconstruction of fine-grained product details in masked regions, complementing the shortcomings of relying solely on MSE loss in the latent space.
In experiments, we demonstrate that our \method can achieve state-of-the-art performance in generating human-product images, especially excelling at preserving high-fidelity details.

In summary, our main contributions are as follows:
\begin{itemize}[leftmargin=10pt,itemsep=0.8ex,topsep=0.5ex]
    \item We propose a novel high-fidelity reference-based inpainting framework , \textbf{\method}, incorporating \textit{Shared Enhancement Attention (SEA)} for enhancing fine-grained product features and \textit{Detail-Aware Loss (DAL)} for providing precise pixel-level supervision.
    
    \item We curate a large-scale and diverse dataset, \textit{HP-Image-40K}, with samples curated from self-synthesis data and processed with automated filtering, to provide a solid foundation for model training.
    
    \item Extensive experiments validate the effectiveness of \method, showing its superior ability to generate detail-preserving human-product images.
\end{itemize}
\section{Related Works}
\label{sec:related}

\subsection{Text-to-Image Generation}
Text-to-image (T2I) generation has experienced significant progress in recent years, making it possible to synthesize images directly from textual input. 
Early methods primarily relied on Generative Adversarial Networks (GANs) \citep{reed2016generative, xu2018attngan, qiao2019mirrorgan}, 
while auto-regressive transformers \citep{ramesh2021zero, yu2022scaling, bai2024meissonic} later demonstrated their potential.
The introduction of diffusion models has revolutionized T2I generation \cite{zhang2025trade,chen2025empirical}, leading to rapid advancements in related applications, such as image customization \citep{ruiz2023dreambooth, zhou2024magictailor,he2026re,lin2024pair,bai2023integrating}, image editing \citep{lin2025jarvisart,lin2025jarvisevo,huang2025dual,zhang2024objectadd,zhang2024artbank,zhang2025u,bai2024humanedit}, consistent image generation \citep{zhou2025identitystory, song2025scenedecorator}, and controllable generation \citep{chen2026posteromni, wang2024diffx, chen2025postercraft, shen2024sg,ma2025lay2story,he2025plangen, lin2024difftv,ling2025mofu,cao2025relactrl}.
However, while T2I diffusion models have achieved remarkable success, generating high-quality human-product images remains a challenging task due to the difficulty of preserving intricate product details. 
In this study, we explore reference-based inpainting to enable precise and seamless integration of humans and products.

\subsection{Image Inpainting}
Image inpainting aims to restore missing or corrupted regions in an image while maintaining visual coherence. Classical methods relied on optimization techniques \cite{bertalmio2000image} or patch-based approaches \cite{criminisi2004region} to fill in gaps based on surrounding context. 
Diffusion models have further advanced the field, offering powerful tools for inpainting by iteratively denoising images from latent representations \cite{saharia2022palette,avrahami2023blended,manukyan2023hd}.
These models can further incorporate additional conditions, providing better control over inpainting tasks \cite{xie2023smartbrush,yang2023uni}. 
Among these, reference-based inpainting \cite{yang2023paint,song2025insert,mao2025ace++,li2025ic} has emerged as a promising paradigm, leveraging reference images to guide the restoration process and ensure consistency in visual context.
Despite these advancements, current reference-based inpainting methods still face limitations and thus achieve suboptimal results when applied to human-product images. 
This highlights the need for a more targeted end-to-end solution to generate detail-preserving human-product images.
\section{Methodology}
\label{sec:method}

\begin{figure*}[t]
    \centering
    \includegraphics[width=1\linewidth]{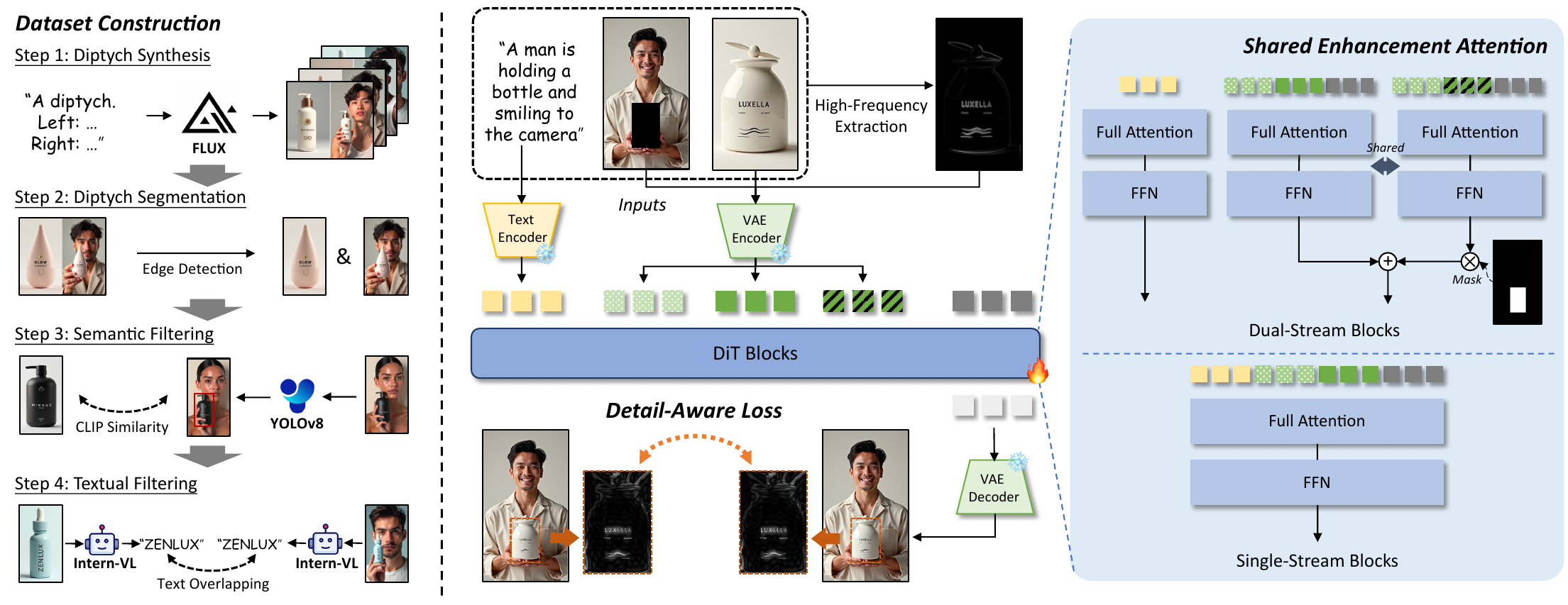}

    \caption{
    \textbf{Overview of \method.}
    HiFi-Inpaint is a high-fidelity reference-based inpainting framework tailored for generating human-product images.
    To support model training, we construct \textit{HP-Image-40K}, a large-scale dataset of human-product images, collected through a self-synthesis pipeline combined with automated filtering to ensure high-quality samples (Sec.~\ref{sec:data}). 
    Furthermore, we introduce two key techniques: (i) \textit{Shared Enhancement Attention (SEA)}, designed to refine fine-grained product features by leveraging high-frequency map tokens within dual-stream visual DiT blocks (Sec.~\ref{sec:model}), and
    (ii) \textit{Detail-Aware Loss (DAL)}, developed to enforce precise pixel-level supervision by utilizing high-frequency information, enabling the reconstruction of intricate product and human details (Sec.~\ref{sec:training}).
    }
    \label{fig:pipeline}

\end{figure*}

\subsection{Overview}
\label{sec:overview}
We propose \textbf{\method}, a high-fidelity reference-based inpainting framework for generating high-quality human-product images.
Given a concise text prompt $T$, a masked human image $\mathbf{I}_h$, and a product reference image $\mathbf{I}_p$, the goal of our \method is to generate an image $\mathbf{I}_g$ that seamlessly integrates the visual content of $\mathbf{I}_p$ into the mask region of $\mathbf{I}_h$ while following the description of $T$.
As illustrated in Fig.~\ref{fig:pipeline}, 
we start by constructing a new dataset called HP-Image-40K, with samples synthesized by a pretrained text-to-image (T2I) model and processed by automatic filtering (Sec.~\ref{sec:data}).
Then, we design a high-frequency map-guided DiT framework that employs a token merging mechanism and integrates Shared Enhancement Attention (SEA) for enhancing product features (Sec.~\ref{sec:model}).
Furthermore, we introduce Detail-Aware Loss (DAL) to achieve high-frequency pixel-level supervision (Sec.~\ref{sec:training}).
Below we delve into the details of the above improvements.

\subsection{Dataset Construction: HP-Image-40K}
\label{sec:data}
Collecting real-world human-product image data is time-consuming and labor-intensive.
% The insufficiency of training data significantly limits the model's generalization for HP image generation.
To liberate model training from data constraints, we use a pretrained T2I model to synthesize desired samples and employ an automatic filtering process to process them, enabling the acquisition of large-scale and diverse data with minimal human intervention.
Finally, we construct a new dataset called \textit{HP-Image-40K}, consisting of 40,000+ high-quality samples to facilitate the training of our model.
The entire procedure is detailed as follows:

\begin{enumerate}[leftmargin=3em,itemsep=0.8ex,topsep=0.5ex]
\item \textbf{Diptych Synthesis:}
We begin by utilizing FLUX.1-Dev \cite{FLUX} to generate diptych-format images, leveraging its capability to retain concept consistency within a generated image \cite{cai2025diffusion}.
We design a dedicated prompt template: ``{A diptych. left: [product description] right: [product and human description]}” to guide the model in producing semantically aligned diptychs, with the left side depicting a product and the right side showing the expected human-product image.

\item \textbf{Diptych Segmentation:} 
Next, we segment the diptychs to extract individual product images and human-product images for subsequent filtering. To simplify the process, we utilize an edge detection algorithm for efficient segmentation. Specifically, a Sobel filter \cite{danielsson1990generalized} is applied to precisely locate the vertical boundary between the two halves, ensuring an accurate separation of the left-side product image and the right-side human-product image.

\item \textbf{Semantic Filtering:} 
Ensuring product consistency across the two segmented images is a crucial step in collecting high-quality data.
For each image pair, we utilize YOLOv8 \cite{yolov8} to localize the product region in the image. We then compute the CLIP \cite{radford2021learning} similarity between the cropped image and the product image.
Only pairs with high similarity scores are selected, ensuring highly consistent samples.

\item \textbf{Textual Filtering:} 
Textual content on products often represents key brand information, labels, or instructions.
To further enhance data quality, we extract textual content from both the product image and the human-product image using InternVL \cite{chen2024internvl}.
We then compare the extracted strings by evaluating their overlapping degree. 
Only pairs with high textual consistency are retained, guaranteeing that the selected samples accurately preserve textual fidelity.

\end{enumerate}

% Through the four-step procedure, we construct the \textit{HP-Image-40K} dataset consists of high-quality samples.
Finally, each sample contains a text prompt $T$, a masked human image $\mathbf{I}_h$, a product image $\mathbf{I}_p$, and a targeted human-product image $\mathbf{I}_{gt}$.
Specifically, $T$ is generated by driving InternVL \cite{chen2024internvl} to describe $\mathbf{I}_t$, and $\mathbf{I}_h$ is produced by masking the region of the detected product.
All samples are carefully curated to ensure high quality, supplementing diverse and robust training data to support the generation of high-fidelity human-product images.

% \begin{figure}[t]
%     \centering

%     \includegraphics[width=1\linewidth]{figure_files/hf.pdf}
%     \vspace{-6mm}
%     \caption
%     {
%         \textbf{Comparison with the Canny algorithm.}
%         While Canny detects all edges, leading to significant background clutter (\textcolor{red}{\textit{red}} frame), the adopted algorithm highlights key elements like text and logos (\textcolor{blue}{\textit{blue}} frame), by being responsive to specific frequencies.
%     }
%     \vspace{-3mm}
    
%     \label{fig:hf}
     
% \end{figure}

\begin{figure}[t]
    \centering
    \begin{minipage}[t]{0.48\linewidth}
        \centering
        \includegraphics[width=\linewidth]{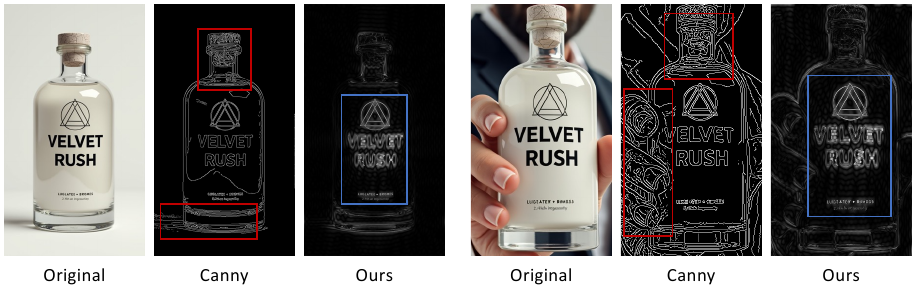}
        \caption
        {
            \textbf{Comparison with the Canny algorithm.}
            While Canny detects all edges, leading to significant background clutter (\textcolor{red}{\textit{red}} frame), the adopted algorithm highlights key elements like text and logos (\textcolor{blue}{\textit{blue}} frame), by being responsive to specific frequencies.
        }
        \label{fig:hf}
    \end{minipage}
    \hfill
    \begin{minipage}[t]{0.48\linewidth}
        \centering
        \includegraphics[width=\linewidth]{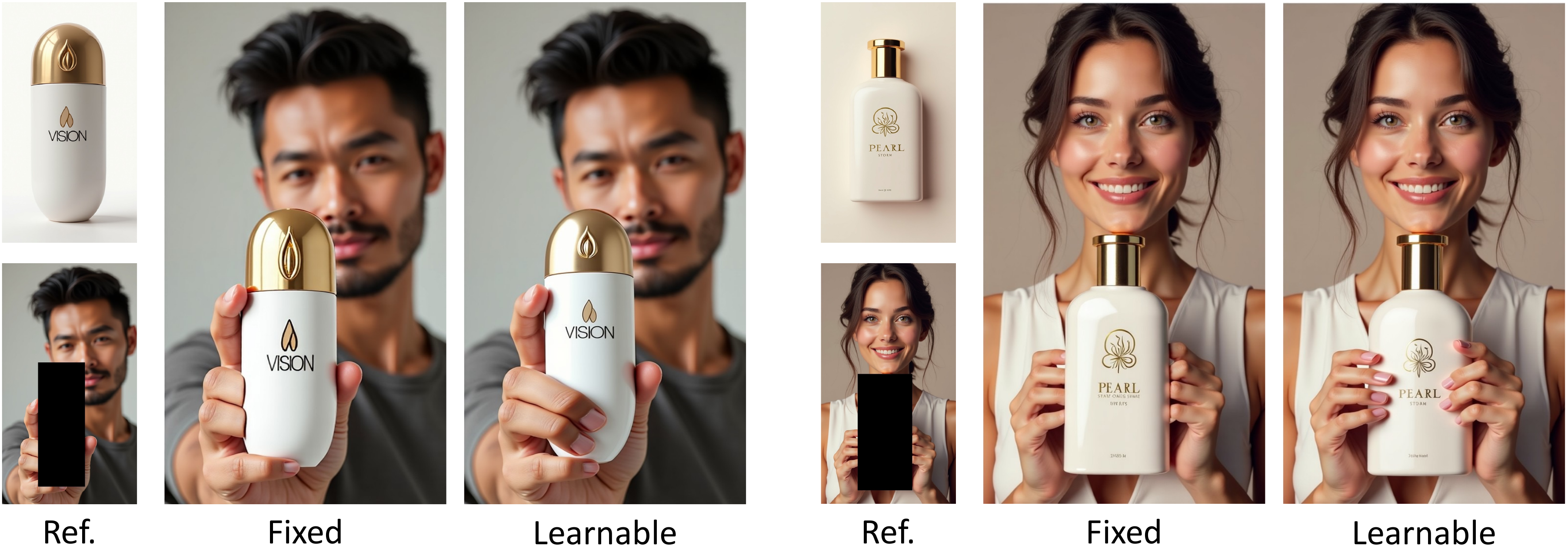}
        \caption
        {
            \textbf{Comparison with fixed weighting of SEA.} 
            Adopting a learnable weighting factor produces more harmonious and realistic results, whereas using a fixed one often leads to visual artifacts and conflicts across the inpainting region.
        }
        \label{fig:alpha}
    \end{minipage}
\end{figure}

\subsection{High-Frequency Map-Guided DiT}
\label{sec:model}

We design a high-frequency map-guided DiT framework based on FLUX.1-Dev \cite{FLUX}.
High-frequency extraction is first adopted, and the token merging mechanism is leveraged to integrate image conditions effectively.
To improve visual fidelity, we introduce \textit{Shared Enhancement Attention (SEA)}, which refines the visual features within masked regions by incorporating high-frequency map tokens into dual-stream visual DiT blocks.

\begin{wrapfigure}{r}{0.6\linewidth}
\vspace{-4mm}
\small
\hrule
\vspace{1mm}
\captionsetup{justification=raggedright, singlelinecheck=false}
\captionof{algorithm}{High-Frequency Extraction}
\vspace{-2.5mm}
\hrule
\vspace{1mm}
\begin{algorithmic}[1]
\Require Input image $\mathbf{I}$
\Ensure High-frequency detail image $\mathbf{I}^{\prime}$
\State Compute DFT: $\mathbf{F} \gets \mathrm{DFT}(\mathbf{I})$
\State Shift zero-frequency to center: $\mathbf{F}_c \gets \mathrm{fftshift}(\mathbf{F})$
\State Construct high-pass mask $\mathbf{M}_h$ (center radius $r$ set to zero)
\State Apply mask: $\mathbf{F}_{h} \gets \mathbf{F}_c \odot \mathbf{M}_h$
\State Inverse shift: $\mathbf{F}_{h}^{-1} \gets \mathrm{ifftshift}(\mathbf{F}_{h})$
\State Inverse DFT: $\mathbf{I}^{\prime} \gets |\mathrm{IDFT}(\mathbf{F}_{h}^{-1})|$
\State \Return $\mathbf{I}^{\prime}$
\end{algorithmic}
\vspace{1mm}
\hrule
\label{algo:hf}
\vspace{-2mm}
\end{wrapfigure}
\myparagraph{High-Frequency Extraction.}
As outlined in Algo.~\ref{algo:hf}, we utilize a dedicated frequency-domain filtering method \cite{jahne2005digital} to extract the high-frequency details of the product. 
Specifically, the input image is transformed to the frequency domain via the Discrete Fourier Transform (DFT), where a high-pass filter is applied by suppressing low-frequency components near the center using a circular mask of radius $r$. The filtered spectrum is then transformed back to the spatial domain using the inverse DFT, and the magnitude response is normalized to obtain the high-frequency map. This process is applied to both $\mathbf{I}_p$ and the masked product region of $\mathbf{I}_{gt}$, exhibiting more targeted results than conventional edge detection in this scenario (Fig.~\ref{fig:hf}).

\myparagraph{Token Merging Mechanism.}
Our base model (\ie, FLUX.1-Dev \cite{FLUX}) adopts the MMDiT architecture. 
During the training of MMDiT, textual tokens are encoded from the text prompt $T$, while noisy visual tokens are obtained by VAE encoding the ground-truth image $\mathbf{I}_{gt}$ with noise added.
To inject the conditions of the masked human image $\mathbf{I}_h$ and the product image $\mathbf{I}_p$, we leverage a token merging mechanism (\ie,  concatenate their encoded tokens with original noisy visual tokens for joint modeling), resulting in joint visual tokens $\mathbf{z}_0$ as
\begin{equation}
\mathbf{z}_0 = \text{Concat} \Big(
\mathcal{E}(\mathbf{I}_h), \mathcal{E}(\mathbf{I}_p), N\big(\mathcal{E}(\mathbf{I}_{gt}), t\big)
\Big),
\end{equation}
where $\text{Concat}(\cdot)$ denotes token concatenation, $\mathcal{E}(\cdot)$ is the VAE image encoder and $N(\cdot, t)$ is the noise addition with a random timestep $t$.
Furthermore, we extract the high-frequency map of the product image $\mathbf{I}_p$, to obtain another token sequence called high-frequency visual tokens $\mathbf{z}_0^\prime$ as
\begin{equation}
\mathbf{z}_0^\prime = \text{Concat} \Big(
\mathcal{E}(\mathbf{I}_h), \mathcal{E}\big(H(\mathbf{I}_p)\big), N\big(\mathcal{E}(\mathbf{I}_{gt}), t\big)
\Big),
\end{equation}
where $H(\cdot)$ denotes the extraction of the high-frequency map (\textit{i.e.,} Algo.~\ref{algo:hf}).
We follow \cite{tan2024ominicontrol} to utilize joint visual tokens $\mathbf{z}_0$ during training, where textual tokens $\mathbf{c}_0$ and noisy visual tokens $\mathbf{z}_0$ are first separately refined by single-stream blocks and then jointly processed by dual-stream blocks. 
% As for high-frequency visual tokens $\mathbf{z}^\prime$, we use them in the proposed attention mechanism.
This mechanism facilitates the model to effectively fuse visual information, enabling reference-based inpainting by capturing the relationships between multiple input images.
As for high-frequency visual tokens $\mathbf{z}^\prime$, we leave the discussion about their usage to the following part.

\myparagraph{Shared Enhancement Attention.}
To improve the model's ability to retain fine-grained product details, such as shapes, patterns, and textures, we propose Shared Enhancement Attention (SEA) that explicitly leverages high-frequency information from the product image to enhance the product-specific visual features.
For each dual-stream visual DiT block, we supplement another branch for high-frequency visual tokens, which leverages the same parameters of the original branch.
Let $B_i(\cdot)$ indicate the $i$-th dual-stream visual DiT block and $\mathbf{z}_i$/$\mathbf{z}_i^\prime$ are the joint/high-frequency visual tokens processed by $B_i(\cdot)$.
Specifically, SEA modifies the naive forward process $\mathbf{z}_i = B_i(\mathbf{z}_{i-1})$ to
\begin{equation}
\mathbf{z}_i = B_i(\mathbf{z}_{i-1}) + \alpha_i \cdot \text{Mask}(B_i(\mathbf{z}_{i-1}^\prime), \textbf{M}_{ds}),
\end{equation}
where $\alpha_i$ is a learnable weighting factor, showing better performance than simply fixing to 1 (Fig.~\ref{fig:alpha}), and $\text{Mask}(\cdot, \textbf{M}_{ds})$ denotes attention masking operation with $\textbf{M}_{ds}$ as the down-sampled masked regions of $\mathbf{I}_h$.
This masking constraint prevents unintended interference from irrelevant regions, ensuring that only the most relevant features are refined. Notably, such a parameter-sharing mechanism can maintain model compactness by introducing only one additional parameter.
The design of SEA enables effective integration of high-frequency details with global contextual information, significantly enhancing the model's ability to capture and preserve intricate product-specific features.

\subsection{Detail-Aware Training Strategy}
\label{sec:training}

To align with and further boost the improvements in the model architecture, we make an enhancement to the training strategy by incorporating high-frequency pixel-level supervision.
This enhancement ensures the accurate preservation of fine-grained details in masked regions while complementing the model's ability to maintain global consistency.

\myparagraph{Detail-Aware Loss.}
To address the limitations of latent-level supervision, which struggles to provide precise guidance for capturing fine-grained details, we propose Detail-Aware Loss (DAL). 
This loss leverages high-frequency pixel-level supervision to complement latent-level objectives, encouraging the accurate reconstruction of intricate product details in masked regions.
Specifically, DAL can be formulated as
\begin{equation}
\mathcal{L}_{\text{DA}} = 
\Big|\Big| H(\hat{\mathbf{I}}_{gt}) \odot \mathbf{M} - H(\mathbf{I}_{gt}) \odot \mathbf{M} \Big|\Big|_2^2,
\end{equation}
where $\hat{\mathbf{I}}_{gt}$ denotes the predicted ground truth, $\mathbf{I}_{gt}$ is the real ground truth, and $\mathbf{M}$ is the original masked region of $\mathbf{I}_h$.
This formulation ensures that the supervision focuses specifically on the high-frequency components of the masked regions, which are captured by $H(\cdot)$.
By emphasizing fine-grained details, DAL effectively guides the model to reconstruct these details that are otherwise challenging to recover through latent-level supervision alone.

\myparagraph{Overall Loss Formulation.}
In addition, we also adopt MSE loss in the latent space to ensure stable global optimization.
This latent-level supervision focuses on preserving the overall semantic and coherence of the generated image, while the pixel-level supervision provided by DAL refines the intricate details within the masked regions.
The full formulation of the overall loss can be expressed as
\begin{equation}
\mathcal{L}_{\text{Overall}} = \mathcal{L}_{\text{MSE}} + \mathcal{L}_{\text{DA}},
\end{equation}
where $\mathcal{L}_{\text{MSE}}$ is the latent-level MSE loss to focus on the reconstruction of clean tokens of $\mathbf{I}_{gt}$.
Finally, the model is trained with $\mathcal{L}_{\text{Overall}}$ using flow matching \cite{lipman2022flow}.
Combining these complementary supervision signals, our training strategy achieves a balanced improvement in both global consistency and local detail fidelity.

\section{Experiments}
\label{sec:exp}

\subsection{Setups}

\myparagraph{Implementation Details.}
In our \method, we adopt FLUX.1-Dev \cite{FLUX} as the base model. 
We utilized an internal dataset consisting of approximately 14,000 samples, which was combined with our curated HP-Image-40K dataset to support model training.
We trained the model with a learning rate of $5 \times 10^{-5}$ and a total batch size of 24 for 10,000 steps. All images were processed at a resolution of $1024 \times 576$ pixels.
For comprehensive evaluation, we split 1,000 samples from HP-Image-40K to assess method performance. 
These samples were exclusively used for evaluation and were not included in the training process.

\myparagraph{Compared Methods.}
We compare our method with four approaches capable of handling reference-based inpainting:
(i) Paint-by-Example \cite{yang2023paint}, which leverages CLIP feature representations to capture the appearance of the reference image and generates matching content in the target region;
(ii) ACE++ \cite{mao2025ace++}, which is an instruction-based approach that integrates multi-modal inputs and employs a two-stage training scheme;
(iii) Insert Anything \cite{song2025insert}, which is a framework using in-context editing and DiT for text-guided image insertion; and 
(iv) FLUX.1-Kontext-Dev (denoted as FLUX-Kontext) \cite{FLUX-Kontext}, which is an image editing model optimized for iterative, precise local and global edits. 
To ensure a fair comparison, all methods are evaluated at a fixed resolution of 1024 × 576 pixels and adhere to the same inference configurations and settings.

\begin{table*}[t]
    \belowrulesep=0pt
    \aboverulesep=0pt
     \caption
     {
         \textbf{Quantitative comparison.}
         The results of automatic metrics demonstrate \method's state-of-the-art performance.
        The best and second-best results are marked in \textbf{bold} and \underline{underlined}.
     }

    \centering
    
    \setlength{\tabcolsep}{3mm}  % 2.3mm
    \renewcommand{\arraystretch}{1.3}
    
    \resizebox{1\linewidth}{!}
    {
    \begin{tabular}{l|c|cccc|cc}
    \specialrule{0.1em}{0pt}{0.5pt}
    \specialrule{0.1em}{0pt}{0pt}
        \multirow{2}[4]{*}{Method\vspace{3mm}} & Text Alignment & \multicolumn{4}{c|}{Visual Consistency} & \multicolumn{2}{c}{Generation Quality} \\
        \cmidrule{2-8}      & CLIP-T$\uparrow$ (\%) & CLIP-I$\uparrow$ (\%) & DINO$\uparrow$ (\%) & SSIM$\uparrow$ (\%) & SSIM-HF$\uparrow$ (\%) & LAION-Aes$\uparrow$ & Q-Align-IQ$\uparrow$ \\
        \midrule 
        Paint-by-Example \cite{yang2023paint} & 31.6  & 69.1  & 63.4  & 54.0  & 34.9  & 4.09  & \underline{4.06} \\
        ACE++ \cite{mao2025ace++} & 34.9  & 93.1  & \underline{90.7}  & 58.3  & 37.2  & 4.18  & 4.00 \\
        Insert Anything \cite{song2025insert} & 35.3  & \underline{94.1}  & 89.8  & \underline{62.1}  & \underline{40.0}  & 4.20  & 3.89 \\
        FLUX-Kontext \cite{FLUX-Kontext} & \textbf{36.6}  & 82.5  & 63.1  & 51.6  & 32.0  & \textbf{4.54} & 3.74 \\
        HiFi-Inpaint (Ours) & \underline{36.1} & \textbf{95.0} & \textbf{91.9} & \textbf{63.4} & \textbf{42.9} & \underline{4.40}  & \textbf{4.36} \\
    \specialrule{0.1em}{0pt}{0.5pt}
    \specialrule{0.1em}{0pt}{0pt}
    \end{tabular}%
    }

    \label{tab:quant_results}
    
\end{table*}

\myparagraph{Evaluation Metrics.}
We evaluate the performance of each method from three perspectives:
\textit{(i) Text Alignment}:
To assess how well generated images align with their text prompts, we compute the average CLIP-T score, which measures the similarity between each generated image and its corresponding text using the CLIP \cite{radford2021learning} model.
\textit{(ii) Visual Consistency}:
We evaluate the visual consistency of generated images using multiple metrics. 
CLIP-I measures the similarity between generated images and reference images in the CLIP feature space, 
while DINO \cite{caron2021emerging} provides another feature-based similarity assessment. 
Additionally, we use SSIM \cite{wang2004image} to quantify structural similarity as we have access to the ground-truth human-product images in the test set.
To specifically evaluate the ability of detail preservation, we introduce SSIM-HF, applying a high-pass filter to generated images before calculating SSIM.
 \textit{(iii) Image Quality}:
We assess the overall quality and aesthetics of the generated images using Q-Align-IQ \cite{wu2023q} and LAION-Aes \cite{laion_aes}. These metrics comprehensively evaluate both the technical and perceptual aspects of image quality, including sharpness, naturalness, and aesthetic appeal.
For a more fine-grained evaluation, all metrics, except for CLIP-T, are calculated based on the generated content of the masked regions and the product reference images.

\subsection{Quantitative Comparison}

We report comprehensive quantitative results across a variety of evaluation metrics in Tab.~\ref{tab:quant_results}, showing \method achieves the overall state-of-the-art performance. 
In terms of text alignment, it achieves a competitive CLIP-T, indicating that the generated images are highly consistent with the input text prompts. 
For visual similarity, our method obtains the top CLIP-I (0.950) and DINO (0.919), demonstrating strong alignment with the product reference images in both global and local features.
Structural similarity is also enhanced, as reflected by the highest SSIM (0.634) and SSIM-HF (0.429), further confirming \method's ability to preserve fine-grained details.
For aesthetic and image quality metrics, our method achieves the best or competitive results on LAION-Aes (4.40) and Q-Align-IQ (4.36), indicating that the outputs are not only faithful to the input but also visually pleasing and of high technical quality.
In contrast, FLUX-Kontext exhibits relatively weak performance, with notably low CLIP-I (0.712) and DINO (0.631), reflecting poor textual and visual consistency. ACE++ and Insert Anything perform moderately well, with Insert Anything showing slightly better detail preservation (e.g., higher SSIM than ACE++), but both are consistently outperformed by our \method, especially in the metrics of visual consistency. Moreover, Paint-by-Example exhibits outdated performance of visual consistency in such a challenging scenario.
In summary, these results highlight that our method delivers overall superior performance in terms of text alignment, visual consistency, and generation quality.

\begin{figure*}[t]
    \centering

    \includegraphics[width=0.95\linewidth]{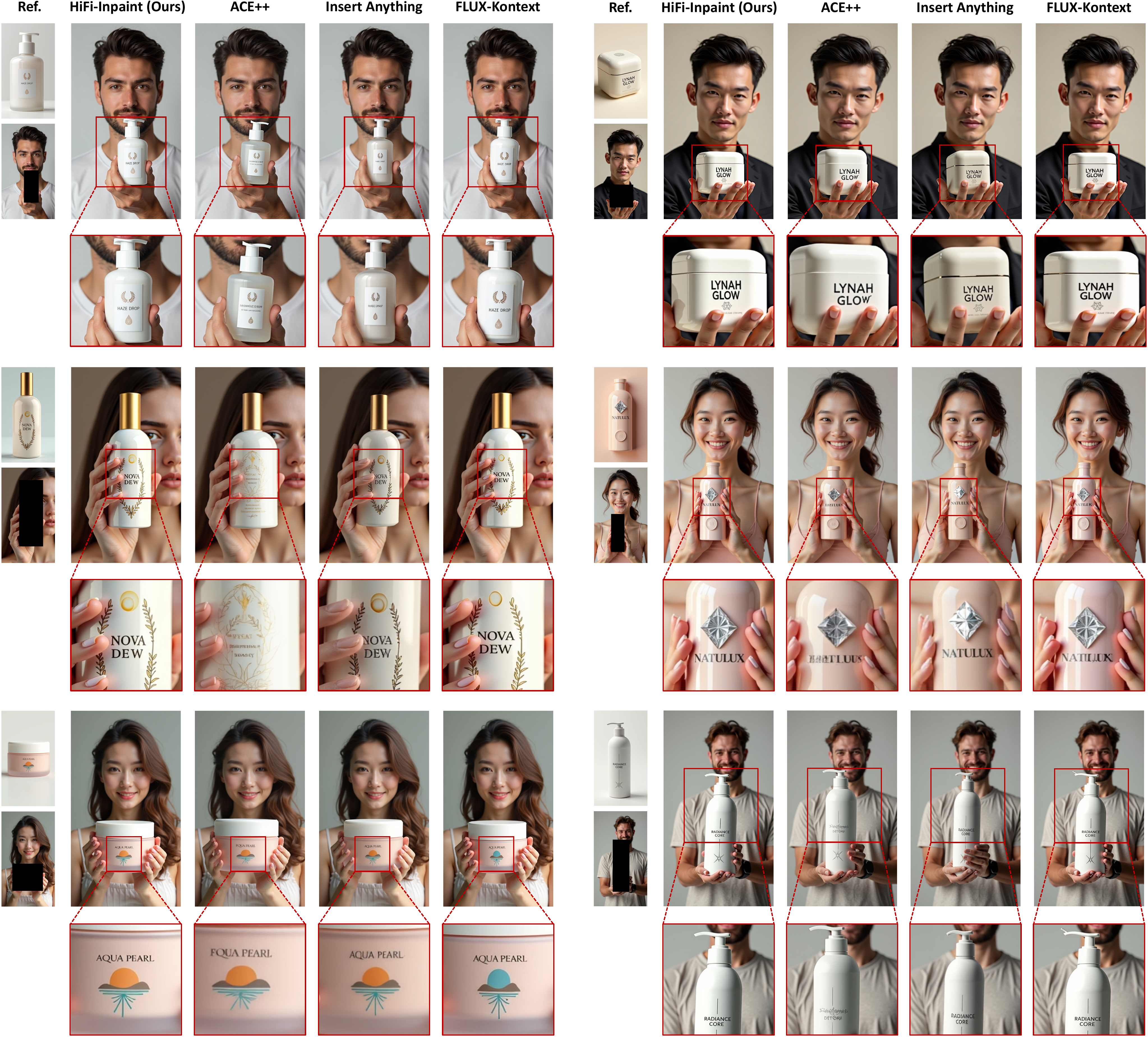}

    \caption
    {
        \textbf{Qualitative comparison.} 
        Compared to existing methods, our \method exhibits remarkable performance in generating high-quality human-product images, enabling high-fidelity preservation of product's fine-grained details.
        Key differences are highlighted with \textcolor{red}{\textit{red}} boxes to emphasize critical improvement areas.
        \textit{Zoom in for better view.}
    }
    
    \label{fig:qual_results}
     
\end{figure*}

\subsection{Qualitative Comparison}
The qualitative comparison, which illustrates the results of the four most competitive methods, is presented in Fig.~\ref{fig:qual_results}.
As we can observe, FLUX-Kontext often fails to perform successful inpainting, often generating a standalone product image instead. 
This is likely because general instruction-based editing is insufficient for the model to establish a meaningful relationship between the reference image and the masked region, leading to confusion regarding the visual elements in the inputs. 
Even in cases where object inpainting is successful, FLUX-Kontext still struggles to preserve fine details, resulting in noticeable inconsistencies in structure and texture.
ACE++ demonstrates a better ability to associate the product with the masked region, effectively preserving the overall product shape and, to some extent, retaining text and patterns from the reference image. 
However, it still faces challenges in reconstructing fine-scale details such as small text or intricate logos.
InsertAnything shows improved performance in detail preservation, maintaining more of the fine-grained details and patterns from the original product. 
Nevertheless, when the mask region is small, it tends to produce artifacts, which significantly degrade the quality of the generated results.
In contrast, our \method achieves superior generation performance.
\method is capable of generating realistic and naturally composited images, with products seamlessly aligned to the background and their fine-grained details, including text, patterns, and branding elements, are faithfully preserved. 
Notably, \method is also able to maintain object shape and text details even when the mask region is small, further demonstrating its robustness for challenging scenarios.

\begin{table*}[t]
    \belowrulesep=0pt
    \aboverulesep=0pt
     \caption
     {
         \textbf{Quantitative ablation analysis.}
         Each component of \method contributes to superior performance.
     }
    
    \centering
    
    \setlength{\tabcolsep}{3.1mm}  % 2.3mm
    \renewcommand{\arraystretch}{1.3}
    
    % \resizebox{0.92\linewidth}{!}
    \resizebox{1\linewidth}{!}
    {
        \begin{tabular}{c|ccc|c|cccc|cc}
        \specialrule{0.1em}{0pt}{0.5pt}
        \specialrule{0.1em}{0pt}{0pt}
\multirow{2}[4]{*}{Scheme\vspace{3mm}} & \multirow{2}[4]{*}{Syn. Data\vspace{3mm}} & \multirow{2}[4]{*}{DAL\vspace{3mm}} & \multirow{2}[4]{*}{SEA\vspace{3mm}} & Text Alignment & \multicolumn{4}{c|}{Visual Consistency} & \multicolumn{2}{c}{Generation Quality} \\
\cmidrule{5-11}      
      &       &       &       & CLIP-T$\uparrow$ (\%) & CLIP-I$\uparrow$ (\%) & DINO$\uparrow$ (\%) & SSIM$\uparrow$ (\%) & SSIM-HF$\uparrow$ (\%) & LAION-Aes$\uparrow$ & Q-Align-IQ$\uparrow$ \\
\midrule
A     &       &       &       & 35.4  & 91.8  & 85.4  & 57.7  & 38.4  & 4.29  & \underline{4.40} \\
B     & $\checkmark$ &       &       & 35.8  & 94.5  & 89.9  & \underline{62.4}  & 41.2  & 4.32  & 4.23 \\
C     & $\checkmark$ & $\checkmark$ &       & \textbf{36.2}  & \underline{94.6}  & \underline{90.7}  & 62.3  & \underline{41.8}  & 4.33  & 4.28 \\
D     &       & $\checkmark$ & $\checkmark$ & 35.9  & 92.2  & 87.6  & 59.8  & 40.3  & \underline{4.34}  & \textbf{4.47} \\
E     & $\checkmark$ & $\checkmark$ & $\checkmark$ & \underline{36.1} & \textbf{95.0} & \textbf{91.9} & \textbf{63.4} & \textbf{42.9} & \textbf{4.40} & 4.36 \\
\specialrule{0.1em}{0pt}{0.5pt}
        \specialrule{0.1em}{0pt}{0pt}
        \end{tabular}%
    }

    \label{tab:ablation}
    
\end{table*}

\subsection{User Study}

\begin{wraptable}{r}{0.5\columnwidth}
    \vspace{-4mm}
    \belowrulesep=0pt
    \aboverulesep=0pt
     \caption
     {
         \textbf{User study.}
          The results on three criteria show that \method outperforms others in human preference.
     }

    \vspace{-2mm}
    
    \centering
    
    \setlength{\tabcolsep}{2mm}  % 2.3mm
    \renewcommand{\arraystretch}{1.3}
    
    \resizebox{1\linewidth}{!}
    {
    \begin{tabular}{l|c|c|c}
    \specialrule{0.1em}{0pt}{0.5pt}
    \specialrule{0.1em}{0pt}{0pt}
        Method & Text Align.$\uparrow$ (\%) & Visual Consis.$\uparrow$  (\%) & Gen. Quality$\uparrow$  (\%) \\
        \midrule
        ACE++  & 20.3  & 19.6  & \underline{22.7} \\
        Insert Anything  & \underline{24.9}  & \underline{21.0}  & 21.6 \\
        FLUX-Kontext  & 18.4  & 17.9  & 16.2 \\
        HiFi-Inpaint (Ours) & \textbf{36.4} & \textbf{41.5} & \textbf{39.5} \\
    \specialrule{0.1em}{0pt}{0.5pt}
    \specialrule{0.1em}{0pt}{0pt}
    \end{tabular}%
    }    

    \vspace{-2mm}
            
    \label{tab:user_study}
    
\end{wraptable}

We conduct a user study to further assess overall generation performance from a human perspective.
Specifically, we design a questionnaire with 11 groups of generated images, each paired with the corresponding text prompt, product reference image, and masked human image. Participants were asked to select the best result in each group based on three criteria: text alignment, visual consistency, and generation quality. A total of 31 valid responses were collected, and the averaged selection rates are presented in Tab.~\ref{tab:user_study}.
The results reveal that our \method consistently outperforms existing methods across all three evaluation dimensions, providing better alignment with human preferences. Additionally, the findings from the user study are highly consistent with the quantitative evaluation of automatic metrics, further validating the effectiveness of our approach in delivering superior image generation quality.

\subsection{Ablation Analysis}

\begin{figure*}[t]
    \centering

    \includegraphics[width=1\linewidth]{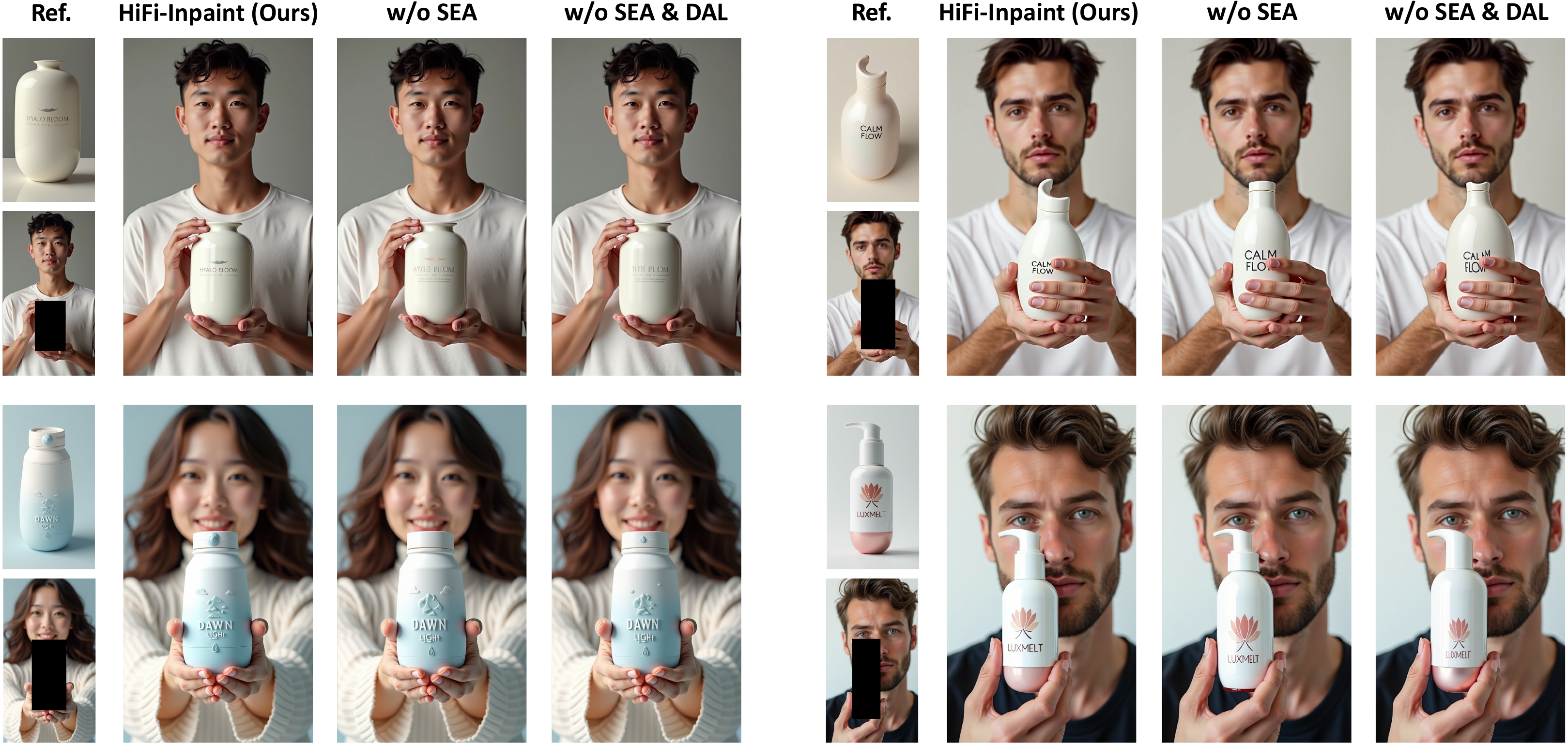}
    
    \caption
    {
        \textbf{Qualitative ablation analysis.} 
        The results demonstrate the effectiveness of both Shared Enhancement Attention (SEA) and Detail-Aware Loss (DAL) in improving the quality of generated human-product images. 
        Our \method, integrating these techniques, achieves the best overall performance with superior detail preservation.
        \textit{Zoom in for better view.}
    }
    
    \label{fig:abl_results}
     
\end{figure*}

To evaluate the effectiveness of each component, we conduct a systematic ablation study by disabling key components while keeping the overall model architecture and training settings.
We report quantitative results for five ablation schemes in Tab.~\ref{tab:ablation}, where Scheme E denotes our final proposed method.
Moreover, we highlight the key ablation comparisons by providing qualitative results in Fig.~\ref{fig:abl_results}.

\myparagraph{Ablation of Synthesized Data.}
We evaluate the improvements brought by the proposed HP-Image-40K dataset, as shown by the comparison of ``Scheme A \& Scheme B” and ``Scheme D \& Scheme E” in Tab.~\ref{tab:ablation}. 
As we can see, under otherwise identical conditions, incorporating our synthetic dataset for model training leads to significant gains in text alignment and visual consistency.

\myparagraph{Ablation of Shared Enhancement Attention.}
We further assess the impact of Shared Enhancement Attention (SEA) by comparing ``Scheme C \& Scheme E. As illustrated in Fig.~\ref{fig:abl_results}, the introduction of SEA enables more precise alignment of intricate details and patterns within the generated images. Furthermore, the quantitative results presented in Tab.~\ref{tab:ablation} show its consistent improvements across multiple metrics, highlighting its comprehensive benefits.

\myparagraph{Ablation of Detail-Aware Loss.}
We assess the contribution of Detail-Aware Loss (DAL) by comparing ``Scheme B \& Scheme C” in Tab.~\ref{tab:ablation}, showing that DAL effectively improves overall performance.
As shown in Fig.~\ref{fig:abl_results}, the model trained without DAL fails to preserve fine-grained elements such as text and intricate patterns, resulting in blurry or semantically incomplete product renderings. These results confirm that DAL is crucial for enhancing the model’s ability to retain critical visual details of products.

\section{Conclusions}
\label{sec:concl}

In this paper, we explore the paradigm of reference-based inpainting for generating high-quality human-product images. We introduce \method, a novel framework that leverages Shared Enhancement Attention (SEA) to capture fine-grained product features and Detail-Aware Loss (DAL) to enable precise pixel-level supervision. To facilitate training, we present HP-Image-40K, a high-quality dataset specifically designed for this task. Experiments demonstrate that our approach achieves superior performance, effectively preserving intricate product details while generating visually coherent images. Future work will focus on enhancing the diversity and realism of the generated images and extending our method to video generation.

\section*{Acknowledgments}
This work was supported in part by the NSFC under Grant No. 62472403, in part by the 2035 Key Research and Development Program of Ningbo City under Grant No. 2024Z123, and in part by the InnoHK initiative of the Innovation and Technology Commission of the Hong Kong Special Administrative Region Government via the Hong Kong Centre for Logistics Robotics. We thank Hao Yang and Ruibiao Lu for their helpful discussions and technical support.

\clearpage

\bibliographystyle{plainnat}
\bibliography{main}

\clearpage
\appendix

\setcounter{secnumdepth}{2}
\setcounter{figure}{6}

% \section{More Details of Methodology}

\section{HP-Image-40K Dataset Statistics}
% 介绍一下数据集的信息（词云，分类啥的），显示一下工作量

To better demonstrate the diversity and comprehensiveness of the HP-Image-40K dataset, we provide detailed statistics in terms of mask area ratio and product categories. These statistics highlight the broad coverage of the dataset, making it a valuable resource for training robust and generalizable models across various real-world scenarios for generating high-quality human-product images.

% \paragraph{Aspect Ratio Distribution.}
% The aspect ratio is defined as the width divided by the height of the bounding region. As shown in Fig.~\ref{fig:aspect_ratio_dist}, the majority of samples fall within the range of $[0.25, 1.10)$, indicating a strong prevalence of vertically oriented or nearly square instances. The most frequent bins are $[0.35, 0.40)$ (7,055 samples), $[0.30, 0.35)$ (5,102), and $[0.40, 0.45)$ (5,923), suggesting that object shapes are generally upright and consistent with common human or product layouts.In addition, our dataset also contains a considerable number of samples with aspect ratios greater than 1.0, which correspond to horizontally elongated (flat) masks. Specifically, approximately 15\% of the masks fall into this category, including a long-tailed distribution of aspect ratios extending up to 6.8. Although these flat-shaped masks are less common than their vertical counterparts, their presence increases the geometric diversity of the dataset. This variety is beneficial for training models that are expected to generalize well across different object shapes and spatial configurations.

\begin{figure}[h]
    \centering
    \begin{minipage}[t]{0.48\linewidth}
        \centering
        \includegraphics[width=\linewidth]{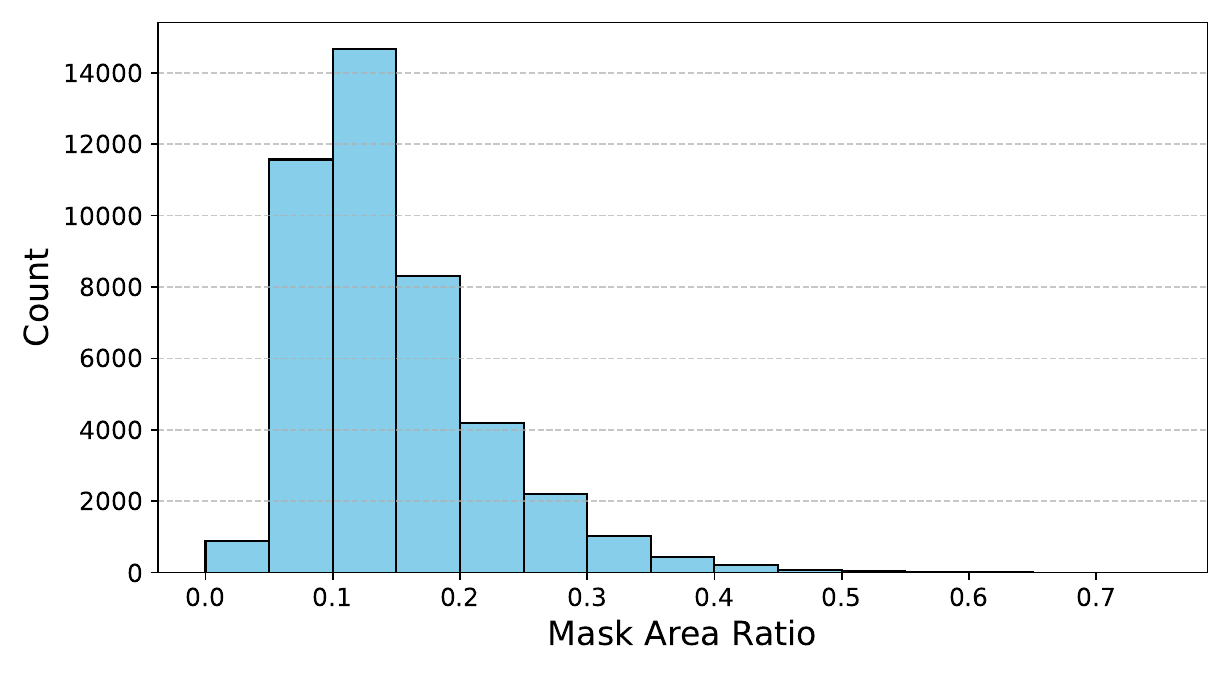}
        \vspace{-5mm}
        \caption{
            \textbf{Histogram of the mask area ratio in HP-Image-40K.} 
            Our HP-Image-40K dataset exhibits a diverse range of mask area ratios, effectively covering various real-world scenarios.
        }
        \label{fig:mask_area_ratio}
    \end{minipage}
    \hfill
    \begin{minipage}[t]{0.48\linewidth}
        \centering
        \includegraphics[width=\linewidth]{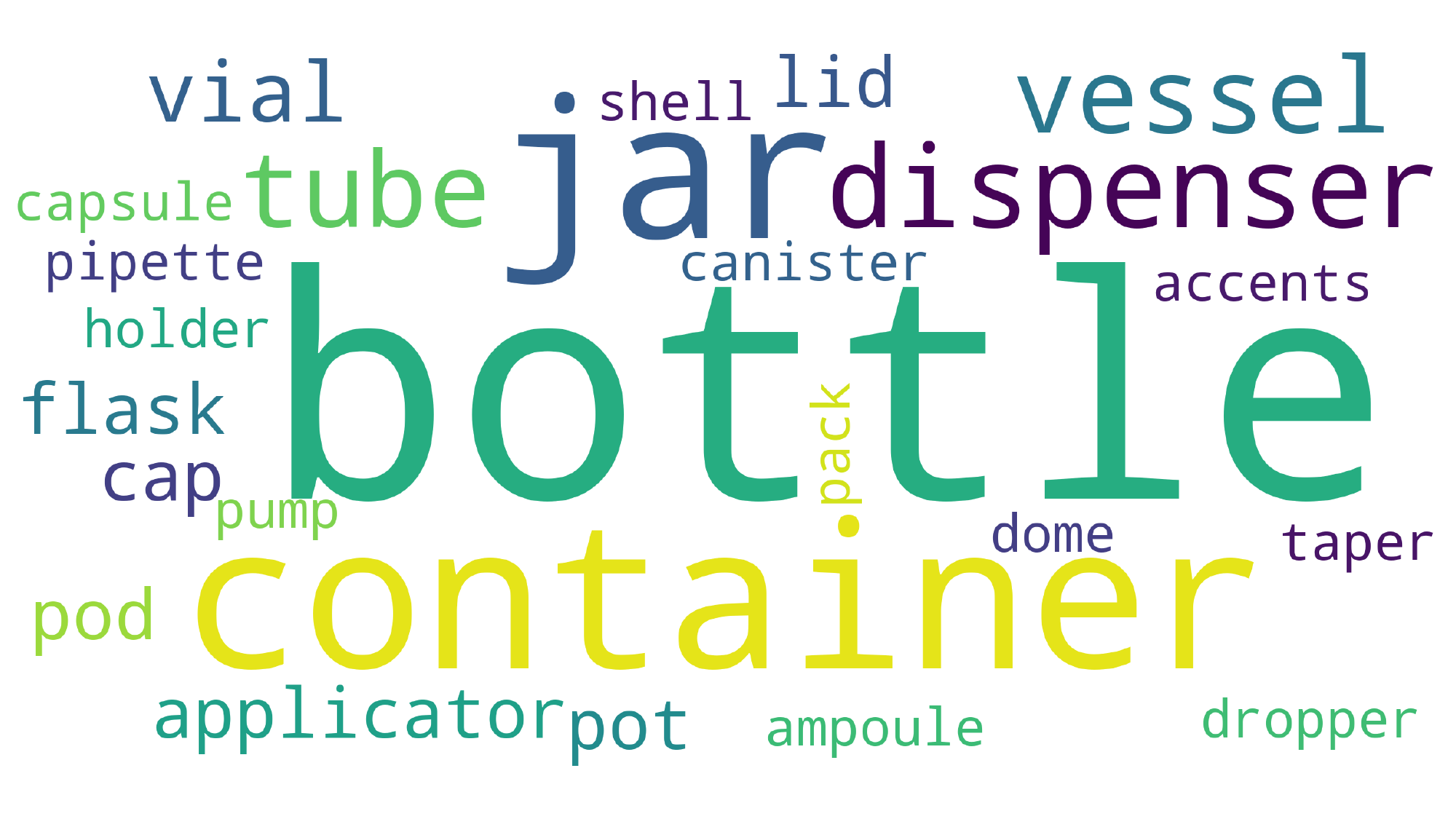}
        \vspace{-5mm}
        \caption{
            \textbf{Word cloud of the product categories in HP-Image-40K.} 
            The dataset encompasses a wide variety of product categories, providing the model with a wide range of shapes, materials, and structures for training.
        }
        \label{fig:wordcloud}
    \end{minipage}
\end{figure}

\myparagraph{Mask Area Ratio.}
The HP-Image-40K dataset includes a wide range of mask area ratios, as shown in Fig.~\ref{fig:mask_area_ratio}. The mask area ratio, defined as the proportion of the mask area to the total image area, varies significantly across the dataset. This variation ensures that the dataset covers diverse object sizes and spatial distributions, from small, localized objects to large, prominent ones. Such mask area ratio diversity is critical for training models that can handle objects of different scales and spatial contexts, improving their performance across various application scenarios.

\myparagraph{Product Categories.}
The HP-Image-40K dataset also features a rich variety of product categories, as visualized in the word cloud in Fig.~\ref{fig:wordcloud}. These categories include bottles, containers, jars, tubes, and dispensers, among others. This diversity not only reflects the dataset's real-world applicability but also enriches the feature representation for model training. By exposing models to a broad spectrum of shapes, materials, and structural characteristics, the dataset enhances the model's ability to generalize across multiple domains and adapt to different product types.

\section{Information of Internal Real-World Dataset}
In addition to the synthetic HP-Image-40K dataset, we further construct a internal real-world dataset collected from publicly available internet images to evaluate model generalization under more realistic conditions. After preprocessing, the real-world dataset is aligned with the synthetic dataset in terms of image resolution and aspect ratio to ensure a fair comparison protocol.

Compared to the synthetic data, the real-world dataset exhibits substantially higher diversity and complexity. It contains a wide range of scenes (indoor and outdoor environments), diverse human subjects with varying poses and interactions, and products with more complex appearances, materials, textures, and branding details. The visual conditions also vary significantly in lighting, viewpoint, occlusion, and background clutter, making the learning problem more challenging than in the controlled synthetic setting.

For training, we utilize approximately 14,000 preprocessed real-world samples. For evaluation, we curate a separate test set consisting of 2,000 preprocessed real-world samples that are not used during training. This split enables a comprehensive assessment of the model’s robustness and generalization capability in complex real-world scenarios.

\section{More Details of Setups}

\noindent\textbf{Training Configuration.}
We set the LoRA scaling factor $\alpha$ to 256, which is equal to the rank. Although our model supports reference images of arbitrary resolutions, we adopt a padding-then-resizing strategy for both training and evaluation to ensure consistent spatial alignment. Specifically, when a reference image does not match the target resolution of $1024 \times 576$, we first pad it to the target aspect ratio while preserving its original content, and then resize it to the fixed resolution. This strategy avoids geometric distortion and maintains structural integrity of the product appearance.

\noindent\textbf{Baseline Adaptation.}
For fair comparison, all baselines are evaluated under the same input resolution ($1024 \times 576$) and identical masked regions. 
\textit{Paint-by-Example}, \textit{ACE++} and \textit{Insert Anything} natively support reference-based inpainting with multi-image inputs. Therefore, we directly follow their official inference protocols and input formatting without additional modification or prompt engineering.
\textit{FLUX-Kontext} is an instruction-based image editing model that does not support explicit multi-image conditioning. To adapt it to our task, we concatenate the product reference image and the masked human image along the width dimension to form a single composite input. Following the official inpainting usage guidelines, we adopt the instruction prompt: ``Change the object in the black square to the product in the left image.'' This enables the model to interpret the left region as the reference product and perform object replacement within the masked area accordingly.

\section{Evaluation on Real-World Data}
In the main paper, experiments were conducted on synthetic data due to their well-controlled alignment with our task definition. 
To further verify the generalizability of the models, we additionally evaluate HiFi-Inpaint and other baselines on an internal real-world test set containing 2,000 diverse human--product samples, which presents significantly more variation in lighting conditions, pose configurations, and product appearance.

\subsection{Quantitative Comparison}

\begin{table*}[t]
    \belowrulesep=0pt
    \aboverulesep=0pt
     \caption
     {
         \textbf{Quantitative comparison on real-world data.}
         The results of automatic metrics demonstrate \method's overall state-of-the-art performance.
        The best and second-best results are marked in \textbf{bold} and \underline{underlined}.
     }

    \vspace{-2mm}
    
    \centering
    
    \setlength{\tabcolsep}{3mm}  % 2.3mm
    \renewcommand{\arraystretch}{1.3}
    
    \resizebox{1\linewidth}{!}
    {
    \begin{tabular}{l|c|cccc|cc}
    \specialrule{0.1em}{0pt}{0.5pt}
    \specialrule{0.1em}{0pt}{0pt}
        \multirow{2}[4]{*}{Method\vspace{3mm}} & Text Alignment & \multicolumn{4}{c|}{Visual Consistency} & \multicolumn{2}{c}{Generation Quality} \\
        \cmidrule{2-8}      & CLIP-T$\uparrow$ (\%) & CLIP-I$\uparrow$ (\%) & DINO$\uparrow$ (\%) & SSIM$\uparrow$ (\%) & SSIM-HF$\uparrow$ (\%) & LAION-Aes$\uparrow$ & Q-Align-IQ$\uparrow$ \\
        \midrule 
    Paint-by-Example \cite{yang2023paint} & 27.1  & 56.2  & 24.3  & 50.8  & 35.7  & \textbf{4.34 } & 2.23  \\
    ACE++ \cite{mao2025ace++} & 28.2  & 80.1  & 74.2  & 53.5  & 36.6  & 3.90  & \underline{3.47}  \\
    Insert Anything \cite{song2025insert} & 28.9  & \underline{83.1}  & \underline{77.5}  & \underline{55.1}  & \underline{37.8}  & 3.95  & \textbf{3.48} \\
    FLUX-Kontext \cite{FLUX} & \underline{29.0}  & 59.9  & 55.7  & 44.6  & 34.3  & \underline{4.30}  & 2.91  \\
    HiFi-Inpaint (Ours) & \textbf{29.7} & \textbf{86.8} & \textbf{79.8} & \textbf{60.5} & \textbf{44.1} & 4.27  & 3.29  \\
    \specialrule{0.1em}{0pt}{0.5pt}
    \specialrule{0.1em}{0pt}{0pt}
    \end{tabular}%
    }    

    \vspace{-1mm}
            
    \label{tab:quant_results_real}
    
\end{table*}

\begin{figure*}[t]
    \centering

    \includegraphics[width=1\linewidth]{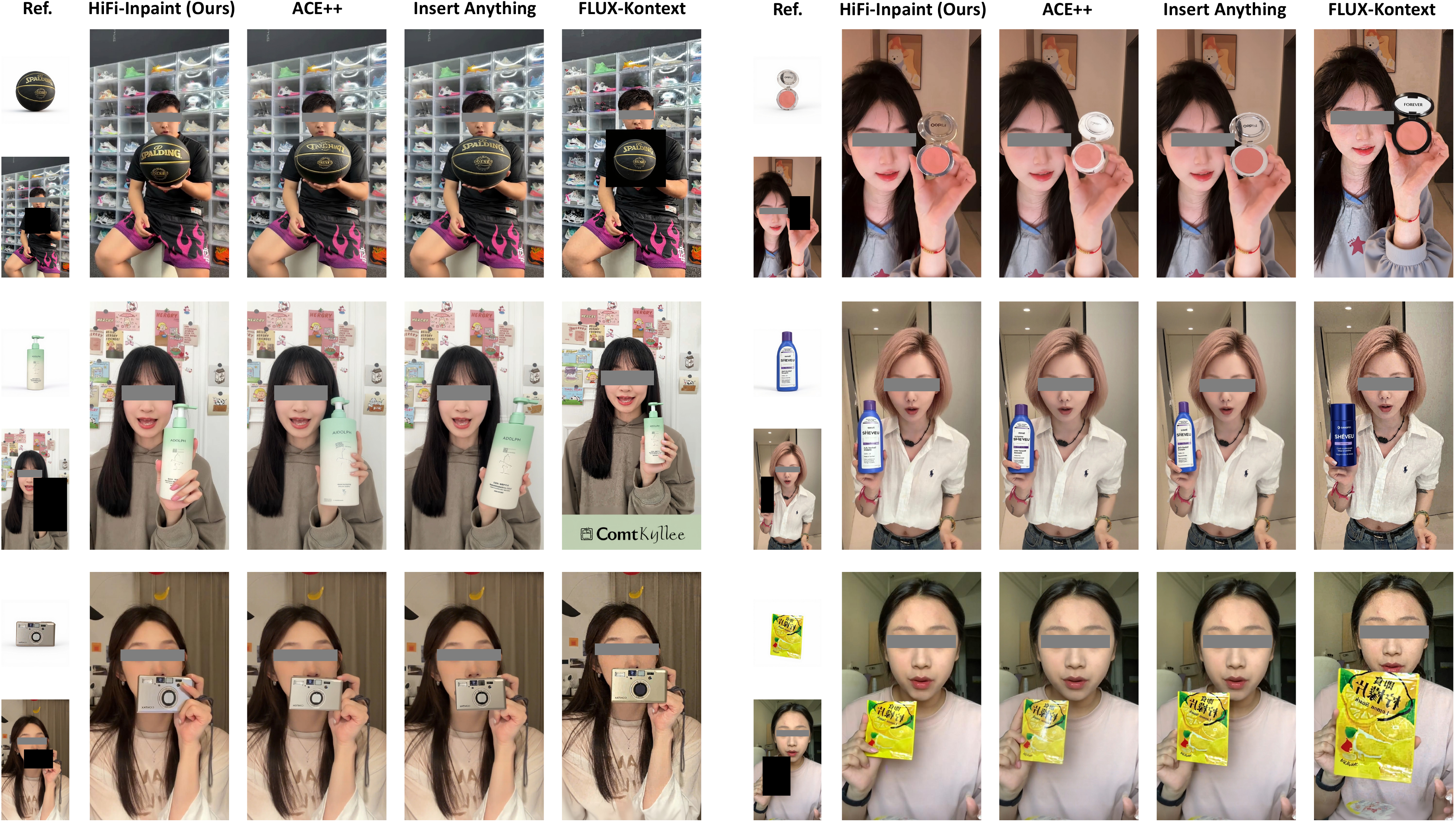}
    \vspace{-4mm}
    \caption
    {
        \textbf{Qualitative comparison on real-world data.} 
        Compared to existing methods, our \method exhibits remarkable performance in generating high-quality human-product images, enabling high-fidelity preservation of fine-grained details.
        The eyes have been obscured to protect the identity of real humans.
        \textit{Zoom in for better view.}
    }
    \vspace{-3mm}
    \label{fig:qual_results_real}
     
\end{figure*}

Tab.~\ref{tab:quant_results_real} reports quantitative comparisons across all metrics, showing that \method remains highly competitive under the more challenging real-world setting. 
For text alignment, \method achieves the best CLIP-T, indicating that the generated content generally stays faithful to textual instructions even when scenes exhibit greater complexity.
In terms of visual similarity, our model attains the highest CLIP-I (86.8) and DINO (79.8), suggesting that \method can better preserve both global product identity and local appearance details when the alignment between reference and target is less constrained than in synthetic cases.
Structural similarity follows a similar trend: \method obtains the top SSIM (60.5) and SSIM-HF (44.1), reflecting strong preservation of object structure and high-frequency characteristics such as text, logos, and fine patterns.
For aesthetic and perceptual quality, \method delivers competitive results, ranking third on both LAION-Aes (4.27) and Q-Align-IQ (3.29). 
Although these scores are slightly lower than the best-performing baselines, they still indicate visually appealing outputs that remain technically coherent with the reference products.
By comparison, FLUX-Kontext exhibits lower CLIP-I (59.9) and DINO (55.7), suggesting that it has more difficulty grounding the reference product under real-world conditions.
ACE++ and Insert Anything achieve moderate to strong performance, with Insert Anything showing relatively good structural and detail preservation, yet both are still outperformed by \method on most visual consistency metrics.
Paint-by-Example lags behind recent methods in this challenging setup, especially on visual consistency and structural similarity.
Overall, the real-world evaluation suggests that \method generalizes well beyond synthetic data and remains robust in preserving product fidelity under realistic variations, while maintaining competitive aesthetic and perceptual quality.

\begin{figure*}[t]
    \centering

    \includegraphics[width=1\linewidth]{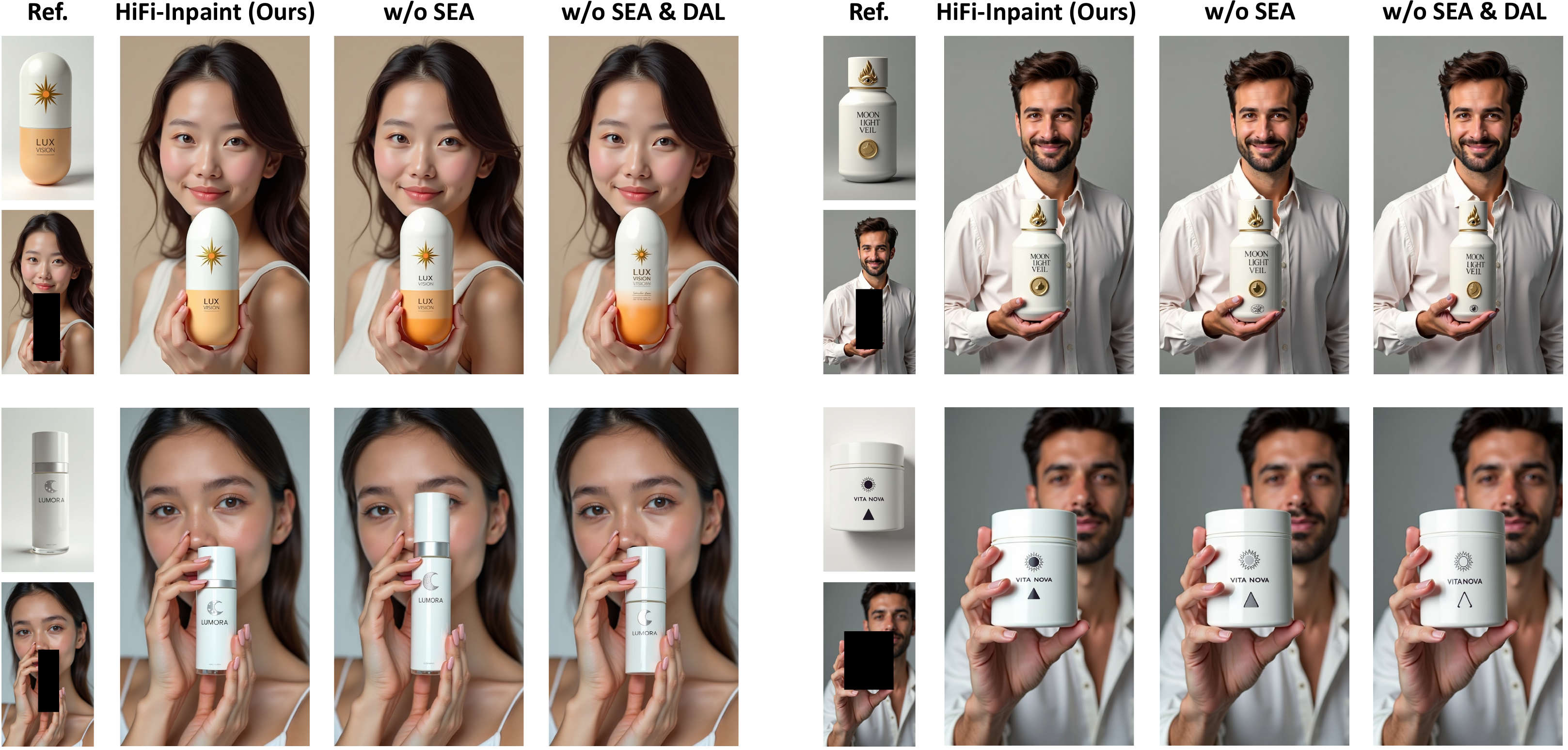}
    \vspace{-5mm}
    \caption
    {
        \textbf{Qualitative ablation analysis.} 
        The results demonstrate the effectiveness of both Shared Enhancement Attention (SEA) and Detail-Aware Loss (DAL) in improving the quality of generated human-product images. 
        Our \method, integrating these techniques, achieves the best overall performance with superior detail preservation.
        \textit{Zoom in for better view.}
    }
    \vspace{-1mm}
    \label{fig:abl_results_more_1}
     
\end{figure*}
\begin{figure*}[t]
    \centering

    \includegraphics[width=1\linewidth]{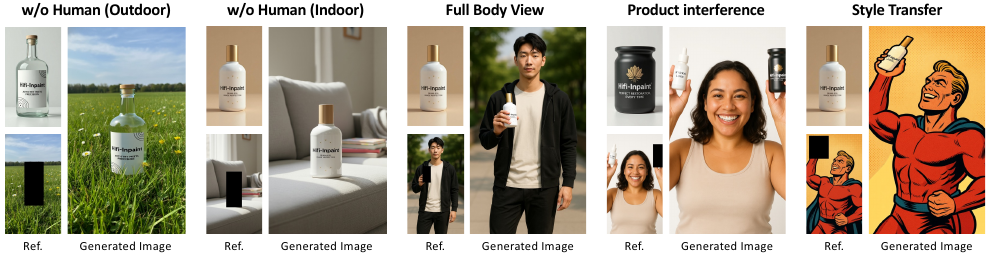}
    \vspace{-5mm}
    \caption
    {
        \textbf{Generalizability analysis of \method.} 
        We further evaluate our \method on several hard cases, demonstrating its potential to generalize to a broader range of scenarios.
        \textit{Zoom in for better view.}
    }
    \vspace{-3mm}
    
    \label{fig:hard_cases}
     
\end{figure*}

\subsection{Qualitative Comparison}

Qualitative comparisons of the four strongest competing methods are provided in Fig.~\ref{fig:qual_results_real}. 
FLUX-Kontext frequently fails to perform correct inpainting, often generating an isolated product instead of integrating it into the masked region. 
This suggests that generic instruction-based editing offers limited capability for grounding the reference product within complex inputs. Even when successful, FLUX-Kontext tends to lose high-frequency details, leading to noticeable inconsistencies in structure and texture.
ACE++ shows a stronger ability to associate the product with the masked region, preserving overall shape and partially retaining textual or patterned elements. However, fine-scale details such as small characters or intricate logos are often not accurately reconstructed.
Insert Anything performs better in detail preservation but tends to introduce artifacts when the masked region becomes smaller, degrading realism and compositional quality.
In contrast, \method produces clean, realistic, and naturally composited results. 
The model faithfully preserves product appearance, including text, patterns, and branding elements, while aligning the inpainted region seamlessly with the surrounding context. 
Importantly, \method remains robust even when the mask is small, maintaining structural integrity and fine-grained details without introducing noticeable artifacts.

\section{Additional Results of Ablation Analysis}
To further validate the effectiveness of key components in \method, we conduct a systematic ablation study, with additional qualitative results shown in Fig.~\ref{fig:abl_results_more_1}.
The examples show that removing individual components leads to noticeable degradation in the detail preservation performance. 
In contrast, the complete \method consistently produces superior results, faithfully preserving critical details and achieving seamless integration with the background.
These results further highlight the contributions of our proposed techniques in tackling challenging inpainting tasks.

\section{Generalizability Analysis}

In Fig.~\ref{fig:hard_cases}, \method is further evaluated on a collection of challenging real-world cases designed to examine the model’s behavior beyond standard inpainting conditions. These examples cover a wide span of difficult scenarios, including images without humans in both outdoor and indoor environments, full-body human views with large pose variations, situations with product interference in the masked image, and cases requiring substantial style adaptation. As illustrated in Fig.~\ref{fig:hard_cases}, \method consistently produces coherent and context-aware completions across these heterogeneous inputs. Even when object scale, lighting conditions, or style distributions deviate significantly from the training distribution, the model is able to integrate the target product naturally into the scene while preserving its key visual attributes. Although certain extreme cases still reveal room for improvement, these results highlight the model’s potential to generalize toward a broader range of practical applications and more diverse deployment conditions.

\end{document}